\documentclass{article}

\usepackage{microtype}
\usepackage{graphicx}
\usepackage{subfigure}
\usepackage{booktabs} 

\usepackage{hyperref}

\usepackage[accepted]{icml2025}

\usepackage{amsmath}
\usepackage{amssymb}
\usepackage{mathtools}
\usepackage{amsthm}
\usepackage{makecell}

\usepackage[capitalize,noabbrev]{cleveref}

\theoremstyle{plain}

\theoremstyle{definition}

\theoremstyle{remark}

\usepackage[textsize=tiny]{todonotes}
\usepackage{pifont}
\usepackage{colortbl} 

\definecolor{wkred}{RGB}{255, 190, 190}
\definecolor{wkblue}{RGB}{210, 230, 250}

\newcommand{\second}{\cellcolor{wkblue}}
\newcommand{\best}{\cellcolor{wkred}}

\icmltitlerunning{Submission and Formatting Instructions for ICML 2025}

\begin{document}

\twocolumn[
\icmltitle{Echo: A Large Language Model with Temporal Episodic Memory}




\begin{icmlauthorlist}
\icmlauthor{WenTao Liu}{11}
\icmlauthor{RuoHua Zhang}{22}
\icmlauthor{Aimin Zhou}{11}
\icmlauthor{Feng Gao}{11}
\icmlauthor{JiaLi Liu}{11}

\end{icmlauthorlist}

\icmlaffiliation{11}{Institute of AI Education, East China Normal University, Shanghai, China}
\icmlaffiliation{22}{School of Computer Science and Technology, East China Normal University, Shanghai, China}


\icmlcorrespondingauthor{Aimin Zhou}{amzhou@cs.ecnu.edu.cn}

\icmlkeywords{Machine Learning, ICML}

\vskip 0.3in
]



\printAffiliationsAndNotice{}  

\begin{abstract}
Research on large language models (LLMs) has shown remarkable performance in domains such as mathematics, programming, and literary creation. However, most studies have focused on semantic memory-based question answering, neglecting LLMs' potential to handle episodic memory (EM)-related queries. This oversight has led to suboptimal performance in applications requiring EM, including emotional companionship, personal AI assistants, and AI teachers.
To address this gap, we introduce Echo, a LLM enhanced with temporal episodic memory. We propose a Multi-Agent Data Generation Framework that guides the model in generating multi-turn, complex scenario episodic memory dialogue data (EM-Train). Temporal information is innovatively incorporated into the LLM training process, and Echo is trained using the EM-Train.
Furthermore, We develop an EM-Test benchmark specifically designed to evaluate LLMs' episodic memory capabilities. The EM-Test assesses performance across various time spans and difficulty levels, providing a comprehensive evaluation of multi-turn episodic memory dialogues.
Our experiments demonstrate that Echo significantly outperforms state-of-the-art LLMs on EM-Test. Additionally, a qualitative analysis reveals Echo's potential to exhibit human-like episodic memory capabilities.
We will open-source all datasets, code, and model weights.
\end{abstract}

\begin{figure}[t!]
  \vspace{-2mm}
    \includegraphics[width=0.5\textwidth]{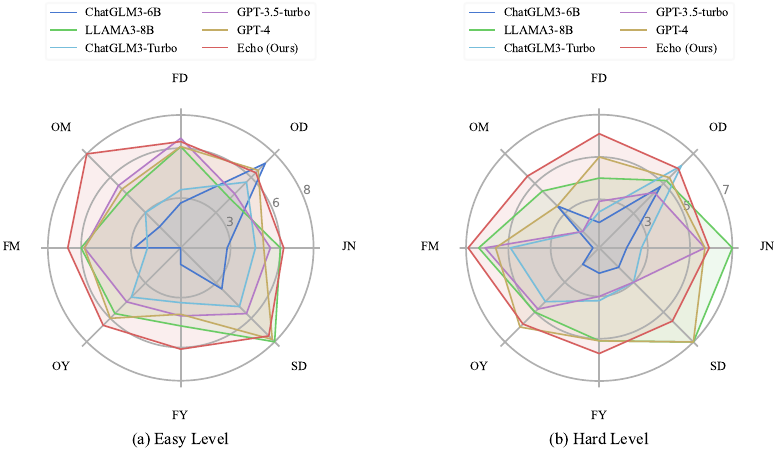}
    \vspace{-6mm}
    \caption{The performance of LLMs across 7 time spans and and two difficulty levels in our EM-Test.}
    \label{fig:performance_llm}
    \vspace{-3mm}
\end{figure}

\section{Introduction}
\label{sec:intro}

Research on large language models (LLMs) has made significant advances in many fields \cite{naveed2023comprehensive,zhao2023survey}, such as mathematical problem \cite{liu2023mathematical}, programming \cite{zhang2023unifying}, and tool usage \cite{qin2024toollearningfoundationmodels}. However, these tasks primarily rely on semantic memory, with little focus on evaluating and enhancing the LLMs' episodic memory capabilities.

\begin{figure}[t!]
\centering

\includegraphics[width=.8\linewidth]{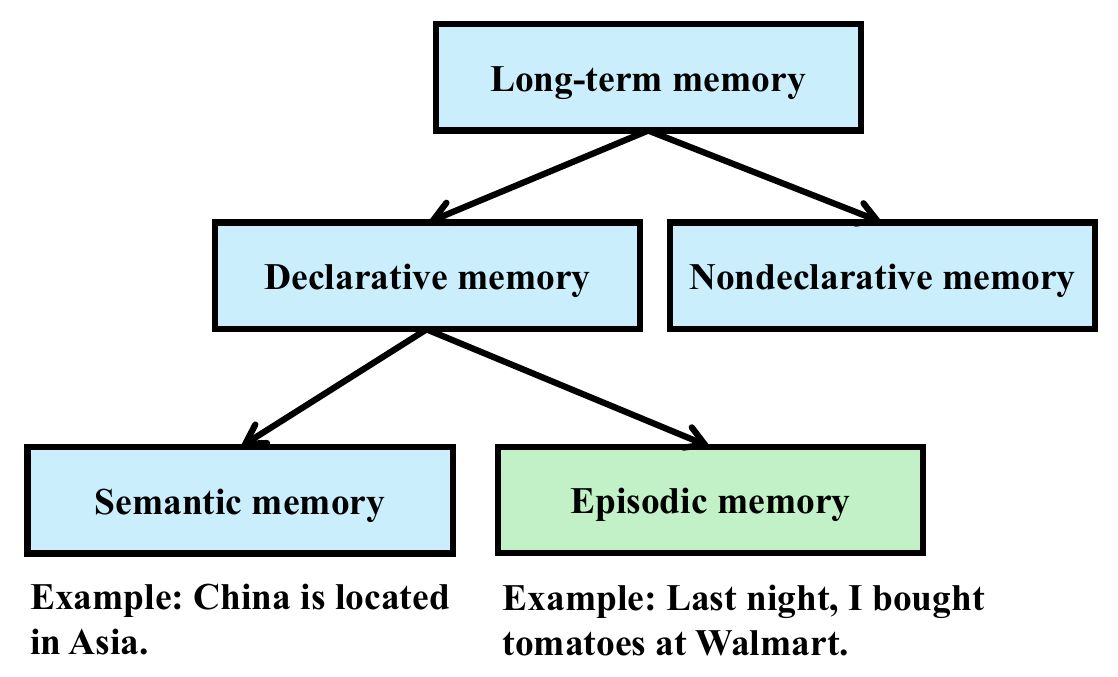} \\
\vspace{1mm}
\caption{The relationship between episodic memory and long-term memory \cite{squire2004memory}.}
\label{fig:em}
\vspace{-5mm}
\end{figure}

Episodic memory is a crucial component of human memory \cite{tulving1983elements,tulving1972episodic}. As shown in Fig.~\ref{fig:em}, long-term memory \cite{squire2004memory} is mainly divided into declarative and non-declarative memory. Declarative memory further comprises semantic memory and episodic memory. Semantic memory involves the recollection of widely accepted concrete facts, which relate to knowledge independent of its context of acquisition \cite{moscovitch2016episodic}. It includes world knowledge, entity memory, language memory, and concept memory, among others. For example, "China is in Asia" or "1+2=3". In contrast, episodic memory refers to time-related event memories centered around the individual, such as "Last night, I bought tomatoes at Walmart".
In fact, episodic memory is not only a fundamental ability of humans but also a critical capability for LLMs, impacting their performance in any multi-turn Q\&A scenarios, such as role-playing \cite{wang2023rolellm}, psychological counseling \cite{ke2024exploring}, and AI teaching \cite{dan2023educhat}. Unfortunately, even the most advanced models (e.g., GPT-4) still perform poorly in terms of episodic memory, often suffering from logical inconsistencies and hallucinations. 

Some methods \cite{zhong2024memorybank,barmann2024episodic,fountas2024human,packer2023memgpt,gao2024memory,hu2023chatdb} have been proposed to enhance the long-term memory capabilities of LLMs. These methods primarily use external storage to retain historical records and design operations to help LLMs retrieve this information for responses. However, these approaches can be time-consuming due to the operations on external storage, and context information may be arbitrarily segmented, leading to information loss. Additionally, these methods do not improve the model's inherent ability to process episodic memory. Episodic memory is thought to be constructive, meaning recall is the (re)construction of a past experience rather than the retrieval of a copy \cite{sprott1933remembering,schacter2012constructive}.

In practice, generative models have an inherent capability to construct and consolidate memories \cite{spens2024generative}. We argue that LLMs face a significant challenge in developing robust episodic memory capabilities due to the limited availability of high-quality episodic memory data. Such data is essential for training models to effectively handle complex, context-dependent interactions.


First, we propose MADGF, a innovative Multi-Agent Data Generation Framework. MADGF simulates and controls multi-turn scenario dialogues between multiple human roles and an AI assistant. The collected dialogue data, named EM-Train, is used to train our Echo model.
In MADGF, three key components are designed: characters, plots, and environments. The design of characters and environments ensures a diverse range of dialogues, while plots guide the LLM to generate dialogue data with enhanced episodic memory capabilities. Additionally, the LLM's training paradigm is modified by incorporating temporal information into each conversation, enriching the temporal background in the interaction process.

Next, we introduce EM-Test, a novel multi-turn dialogue benchmark designed to evaluate episodic memory capabilities. Each instance in EM-Test may contain multiple evaluation points, requiring the model not only to process long-context text effectively but also to recall, reason, and cognitively handle episodic memory information. Each evaluation point is tagged with both time and difficulty levels, enabling a comprehensive assessment.
To reduce manual evaluation efforts, we propose an approach that assesses model performance based on semantic similarity. The feasibility and effectiveness of approach is validated by its strong correlation with human evaluations.

Finally, we conducted both quantitative and qualitative experiments. The quantitative results show that Echo significantly outperforms state-of-the-art LLMs on the EM-Test. Additionally, the qualitative analysis reveals Echo's potential to exhibit human-like episodic memory capabilities.

\section{Related Work}
\label{sec:background}

\paragraph{Methods for Enhancing Long-Term Memory Capabilities}

Some methods have been proposed to enhance the long-term memory capabilities of large models, such as MemoryBank \cite{zhong2024memorybank}, H-EMV \cite{barmann2024episodic}, EM-LLM \cite{fountas2024human}, MemGPT \cite{packer2023memgpt}, MS \cite{gao2024memory}, and CHATDB \cite{hu2023chatdb}. These methods use external storage to retain historical information and design various operations to help LLMs utilize information.

MemoryBank \cite{zhong2024memorybank} introduces a novel memory mechanism specifically designed for LLM. This mechanism processes historical conversation to extract summary information and user portrait. When a user poses a question, the mechanism retrieves relevant information based on similarity and combines it with the summary information and user portrait, to form a Meta Prompt that assists the model in generating responses.
EM-LLM \cite{fountas2024human} adopts a similar method by incorporating key information into preceding prompts. This method effectively handles nearly unlimited context lengths while maintaining high computational efficiency.
MemGPT \cite{packer2023memgpt} enables LLMs to perform tasks beyond the current context limits by simulating extended virtual memory through paging between physical memory and disk storage, akin to how operating systems manage memory to extend LLM context.
MS \cite{gao2024memory}, H-EMV \cite{barmann2024episodic}, and CHATDB \cite{hu2023chatdb} introduce distinct data structures designed for the storage of historical information: namely, a memory-sharing framework, a tree-based storage structure, and a specialized database, respectively. Each of these architectures facilitates the retrieval of pertinent historical data to support the response generation.

These methods require various operations on external storage that can be time-consuming. Moreover, they primarily focus on retrieving a copy of the data, rather than implementing the constructive nature of episodic memory \cite{sprott1933remembering,schacter2012constructive}, failing to enhance the model's inherent ability to process episodic memory.

\paragraph{Methods for Data Generation utilizing LLM}
Manually annotated data is expensive, so many methods \cite{xu2023wizardlm,luo2023wizardmath,zhao2024wildchat,wang2022self,ding2023enhancing,li2023camel} have been proposed to automate data generation utilizing LLMs. Besides obtaining data through user interactions on online platforms using ChatGPT, like WILDCHAT \cite{zhao2024wildchat}, Self-Instruct \cite{wang2022self} was one of the first to propose generating instructions, inputs, and outputs using LLMs to build instruction fine-tuning data. To increase the diversity of instructions, WizardLM \cite{xu2023wizardlm} introduced an evolutionary instruction approach starting from a small set of seed instructions to generate more complex and diverse instruction. Further, WizardMath \cite{luo2023wizardmath} incorporated a reward model to select better instruction data from multiple outputs, collecting higher-quality generated data. Additionally, some methods \cite{ding2023enhancing,li2023camel} propose having LLMs play the roles of both AI assistant and user to collect data, which allows for the collection of multi-turn dialogues. UltraChat \cite{ding2023enhancing} uses this approach to extract instruction data covering various tasks, such as Questions about the World and Creation and Generation. In contrast, CAMEL \cite{li2023camel} focuses on generating instruction data for specific tasks, such as "Develop a trading bot for the stock market."

These LLM-based data generation methods primarily focus on extracting high-quality instruction fine-tuning data grounded in semantic memory from LLMs. In contrast, our MADGF mainly aims to simulate real-life scenarios to generate dialogue content rich in episodic memory.



\section{Mutil-Agent Data Generation Framework}
\label{sec:approach}

The purpose of the Multi-Agent Data Generation Framework (MADGF) is to design multiple human characters interacting with an AI assistant. Through simulating daily conversations, a large multi-turn dialogue dataset enriched with episodic memories is collected for the training of the Echo model. To enhance the diversity and effectiveness of the conversation content, we initially devised three key elements: characters, plots, and environments. Extensive character cards, plots, and temporal information were then generated. Subsequently, we formulated a data generation process that utilizes this information to guide the LLM in producing high-quality episodic memory data (EM-Train).

\subsection{Characters, plots, and Environments}
\label{sec:three key}
\begin{figure}[t!]
\centering

\includegraphics[width=.85\linewidth]{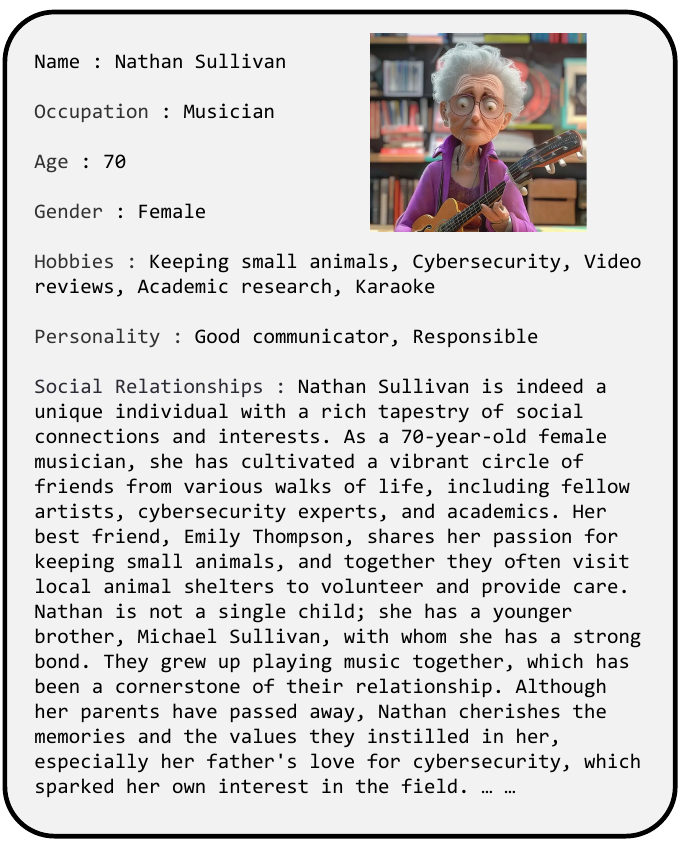} \\

\caption{Example of character card.}
\label{fig:character}
\end{figure}

\paragraph{Characters}
As illustrated in Figure \ref{fig:character}, the design of character cards encompasses seven attributes: "Name," "Occupation," "Age," "Gender," "Hobbies," "Personality," and "Social Relationships." Specifically, we randomly generated attribute values for all attributes except for "Social Relationships." Subsequently, we utilized the LLM to generate the "Social Relationships" attribute values based on the other six attributes.

\paragraph{Plots}
The plots generated by LLMs differ significantly from actual real-life scenarios. Therefore, we manually created an event library, from which 20 events are sampled to form a plot. The library contains three types of events: common events, real events, and hallucinatory events. Common events are designed to enable the model to generate data based on semantic memory questions and answers while enriching the context. They include routine occurrences in daily life, such as greetings, inquiries about common knowledge, and discussions about career-related issues. Real events are events that have actually occurred and are related to episodic memory. They are used to prompt the human role to ask the Echo assistant if it remembers a related event. Hallucinatory events are fabricated events that have never occurred. They are used to prompt the human agent to ask the AI assistant about non-existent events and simultaneously remind the AI assistant not to be misled. Notably, since all event prompts are removed during the training of the Echo model, hallucinatory events help reduce the LLM's tendency to generate false information and enhance the model's understanding and reasoning abilities regarding episodic memory.

\paragraph{Environments}
In the design of environments, we initially considered only temporal information. We first established a series of time-stamped nodes arranged in chronological order (e.g., Monday, September 4, 2006, 21:42:56, Monday, September 4, 2006, 21:55:38). These time-stamped nodes are then automatically added to the conversation history between the human role and the Echo assistant, indicating the time at which each round of dialogue takes place.

\subsection{Data generation process}
\label{sec:dgp}

\paragraph{Prompt Design}

\begin{figure}[t!]
\centering

\includegraphics[width=.8\linewidth]{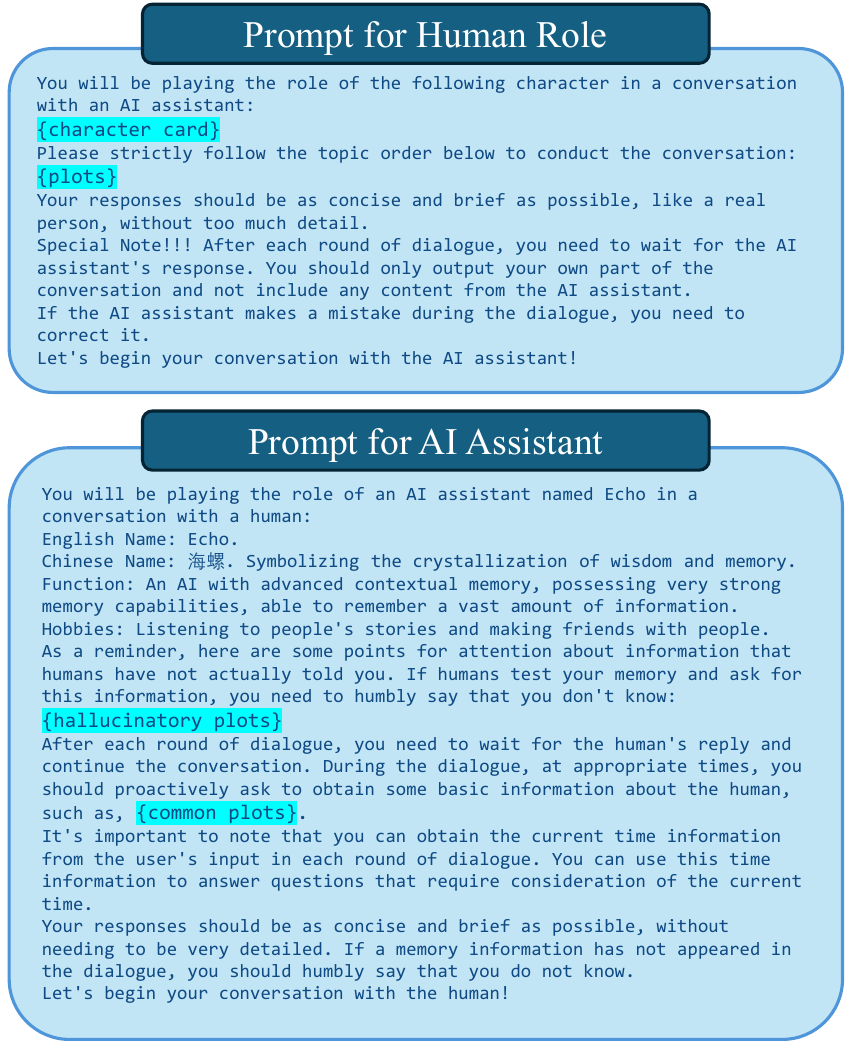} \\

\caption{Prompt template in data generation process.}
\label{fig:template}
\end{figure}

As illustrated in Figure \ref{fig:template}, we designed distinct prompt templates for both the human role and the Echo assistant. The highlighted sections in the figure are replaced with information from Section \ref{sec:three key}. Specifically:

\begin{itemize}
    \item \textbf{Human Role Prompt}: This includes the character card and all plot details, enabling the LLM to assume various human roles and engage in dialogues with the AI assistant according to different plots.
    
    \item \textbf{AI Assistant Prompt}: This incorporates both hallucinatory plots and common plots. This setup helps the LLM acting as the AI assistant to reduce episodic memory hallucinations and proactively seek relevant information in a human-like manner.
\end{itemize}

Based on these prompt templates, we generate initial prompts for the human and the Echo assistant, denoted as $P_u$ and $P_a$, respectively.

\paragraph{The Pseudocode of Data Generation Process}

Algorithm \ref{alg:data_generation} provides the pseudocode for the data generation process. We initialize and maintain two separate history records, $H_u$ and $H_a$, for the human role and the AI assistant using initial prompts $P_u$ and $P_a$, respectively. In lines 4-12 of Algorithm \ref{alg:data_generation}, we alternately control the two agents representing the human and the assistant to engage in dialogue. Temporal information is incorporated during the conversation in lines 6-8. We check if farewell phrases such as "goodbye" or "talk to you later" appear in the response. If any of these phrases are detected, or if the number of conversation rounds exceeds 60, the stopping criterion is considered to be met, and the current data generation process is terminated. Finally, we remove the initial prompt $P_a$ from $H_a$ to obtain the final dataset, denoted as $Data$, which constitutes one piece of data in our EM-Train dataset.

\begin{algorithm}[ht]
    \caption{Pseudocode of Data Generation Process}
    \label{alg:data_generation}
    \begin{algorithmic}[1]
        \REQUIRE Initial prompts $P_u$ and $P_a$
        \ENSURE $Data$
        \STATE Initialize $H_u$, $H_a$
        \STATE $H_u \leftarrow H_u \cup P_u$
        \STATE $H_a \leftarrow H_a \cup P_a$
        \WHILE{stopping criterion not met}
            \STATE $answer_u \leftarrow LLM(H_u)$
            \STATE $time \leftarrow RandomNextTime(time)$
            \STATE $H_u \leftarrow H_u \cup answer_u \cup time$
            \STATE $H_a \leftarrow H_a \cup answer_u \cup time$
            \STATE $answer_a \leftarrow LLM(H_a)$
            \STATE $H_u \leftarrow H_u \cup answer_a$
            \STATE $H_a \leftarrow H_a \cup answer_a$
        \ENDWHILE
        \STATE $Data \leftarrow H_a \setminus P_a$
    \end{algorithmic}
\end{algorithm}

\section{Dataset}

\subsection{EM-Train and Training Paradigm}
Based on MADGF in Section \ref{sec:approach}, we collected and created EM-Train. It consists of 15, 533 data entries, with an average of 16.75 conversation rounds per data entry and an average length of 8, 597 characters. Then, we trained the Echo model using the EM-Train dataset.

\begin{figure}[h]
\centering

\includegraphics[width=.9\linewidth]{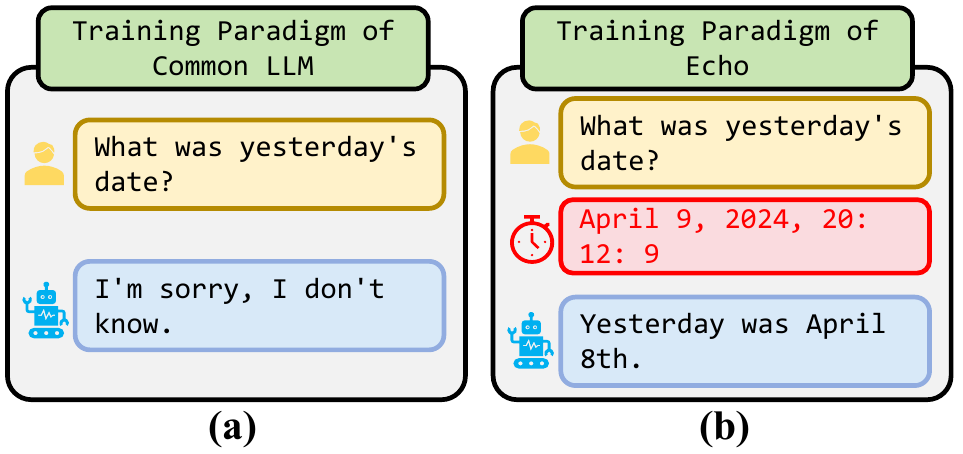} \\

\caption{Illustration of training paradigm changes.}
\label{fig:paradigm}
\vspace{-1mm}
\end{figure}

Compared to the conventional LLM training paradigm (user-assistant), we modified the training paradigm to user-time-assistant. As shown in Figure \ref{fig:paradigm} (a), in traditional LLMs, the chat template for instruction fine-tuning alternates between two roles: user and assistant. In our modified approach, as highlighted in red in Figure \ref{fig:paradigm} (b), we introduced an additional role, "observation," which includes temporal information. During training, the content of the observation does not participate in gradient updates, and the attention mask remains consistent with the traditional decoder-only method. During inference, whenever a user inputs a prompt, real-time information is automatically integrated into the context, enabling the creation of a time-aware AI assistant.




\begin{figure}[h]
\centering

\includegraphics[width=.9\linewidth]{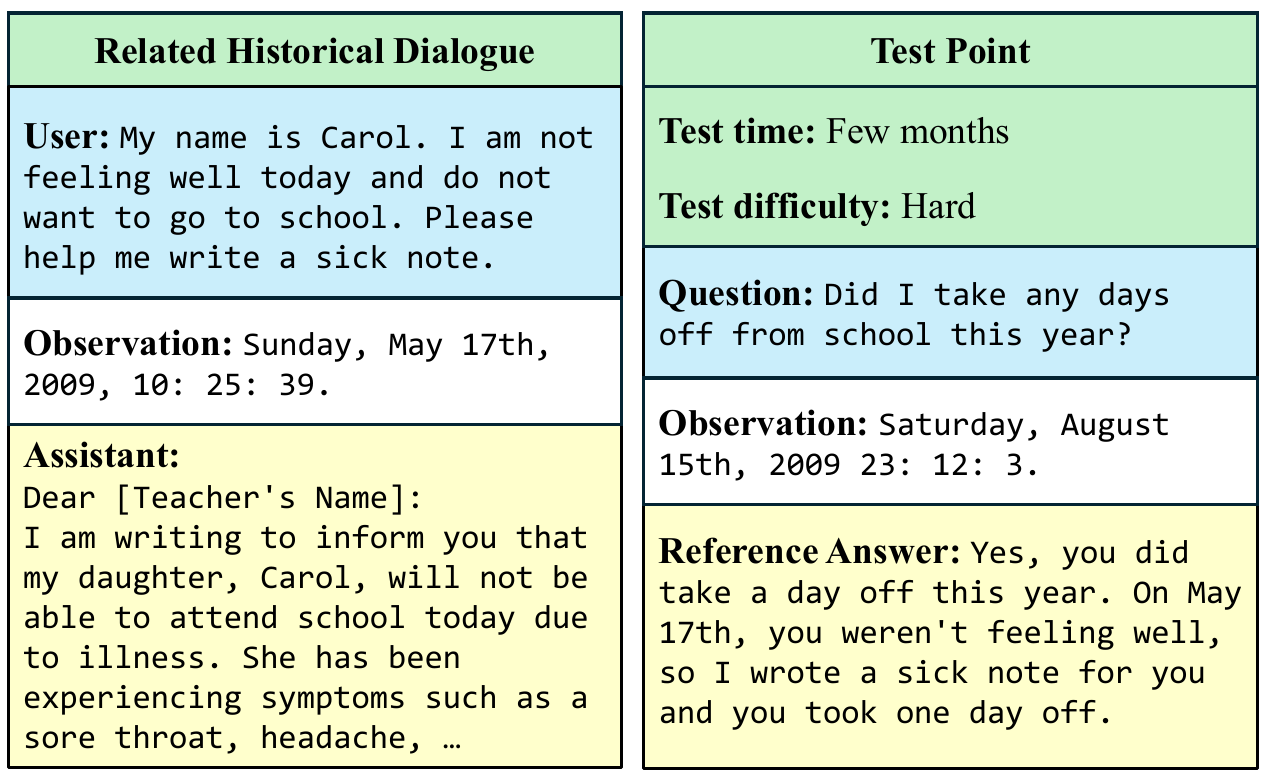} \\

\caption{Example of hard-level test point in EM-Test.}
\label{fig:em_test}
\vspace{-8mm}
\end{figure}

\subsection{EM-Test}

We manually developed a benchmark called EM-Test for evaluating the episodic memory of LLMs. Each test instance consists of multi-turn dialogues. In addition to dialogues that are not directly related to episodic memory testing, these dialogues may include multiple related historical dialogues and a corresponding test point, as shown in Figure \ref{fig:em_test}. At each test point, we annotate the test question (Question), the temporal context (Observation), and the reference answer (Reference Answer).

During testing, we provide all historical dialogues as conversation history to the model. Then, we input the test question and the temporal context to obtain the model's output. The model's output is either manually scored or compared against the reference answer to quantitatively evaluate the episodic memory capabilities of the LLMs.

We labeled the time span and difficulty of the test points to achieve more granular results. In terms of time span, we categorized them into eight types based on the required duration of episodic memory for answering questions: "just now," "one day," "few days," "one month," "few months," "one year," "few years," and "several decades." We also divided the difficulty of test points into easy and hard levels. For an easy-level test point, the model only needs to recall a simple scenario. For a hard-level test point, the model must possess complex episodic memory capabilities. Figure \ref{fig:em_test} provides an example of a hard-level test point. In this example, to answer “Did I take any days off from school this year?”, the model needs to recall writing a leave note for the user and the timing of that event being in the same year as the current one, to deduce the correct answer.

Additionally, we manually created the EM-Test-Without-Time scenario test set to evaluate the model's episodic memory ability without considering time information. Compared to EM-Test, EM-Test-Without-Time does not include temporal context and only considers easy and hard difficulty levels. Table \ref{tab:em_test} presents the relevant statistical information for both EM-Test and EM-Test-Without-Time.


\begin{table}[t!]
\centering
\scriptsize
\caption{Detailed statistics of EM-Test.}
\label{tab:em_test}
\vskip 0.15in
\vspace{-2mm}
\setlength{\tabcolsep}{1.1mm}
\begin{tabular}{llll}
\toprule
\textbf{Time Span} & \textbf{Easy}  & \textbf{Hard} & \textbf{Total} \\
\midrule
\multicolumn{4}{c}{\textbf{EM-Test}}\\
\midrule
just now & 18 & 7 & 25 \\
one day & 5 & 5 & 10 \\
few days & 10 & 8 & 18 \\
one month & 4 & 4 & 8 \\
few months & 4 & 7 & 11 \\
one year & 5 & 4 & 9 \\
few years & 7 & 9 & 16 \\
several decades & 4 & 5 & 9 \\
Overall Number & 57 & 49 & 106 \\
\midrule
\multicolumn{4}{c}{\textbf{EM-Test-Without-Time}}\\
\midrule
Overall Number & 89 & 34 & 123 \\
\bottomrule
\end{tabular}
\vspace{-2mm}
\end{table}
\section{Experiment}
\label{sec:experiment}
\subsection{Experimental Setups}

\paragraph{Selected LLMs.} We evaluate a series of LLMs on EM-Test, including the current state-of-the-art open-source and closed-source models. Particularly, we select LLAMA3-8b \cite{dubey2024llama}, ChatGLM3-6B \cite{glm2024chatglm} for open-source models, and for closed-source models, we employ GPT-3.5-turbo \cite{openai2023gpt4}, GPT-4 \cite{openai2023gpt4}, ChatGLM3-trubo \cite{glm2024chatglm}.

\paragraph{Implementation Details.}
In MADGF, the LLM serving as the agent is Qwen2-72B-Instruct \cite{yang2024qwen2}, which is a high-performance open-source LLM. We use chatglm3-6B \cite{glm2024chatglm} as the base model for Echo, and implement it with full fine-tuning.

\paragraph{Evaluation Methods and Metrics.}

In the quantitative analysis, we first collect the responses of LLMs at the test points, then ask human annotators to score these responses on a scale of 1 to 10. Additionally, we use the widely adopted Sentence Transformer model, all-MiniLM-L6-v2\footnote{\tiny \url{https://huggingface.co/sentence-transformers/all-MiniLM-L6-v2}}, to encode the LLMs' responses and the reference standard outputs provided by the test set, obtaining \(\mathcal{E}_{\text{LLM}}\) and \(\mathcal{E}_{\text{Standard}}\). We then calculate the cosine similarity \(\mathcal{S}\) using the Equation \ref{equ:simi}.

\begin{equation} \label{equ:simi}
    \mathcal{S}  = cos\_sim(\mathcal{E}_{\text{LLM}},\mathcal{E}_{\text{Standard}}) \times 100
\end{equation}

We consider using the Pearson correlation coefficient, denoted by $ \mathcal{R} $, to measure the correlation between the human score and the similarity metric. It is calculated using Equation \ref{equ:pr}.



\begin{equation} \label{equ:pr}
\mathcal{R} = \frac{\sum_{i=1}^{n} (x_i - \bar{x})(y_i - \bar{y})}{\sqrt{\sum_{i=1}^{n} (x_i - \bar{x})^2} \sqrt{\sum_{i=1}^{n} (y_i - \bar{y})^2}}
\end{equation}
where:
\begin{itemize}
    \item \( x_i \) and \( y_i \) are individual sample points indexed with \( i \),
    \item \( \bar{x} \) and \( \bar{y} \) are the mean values of \( x \) and \( y \) respectively,
    \item \( n \) is the number of sample points.
\end{itemize}

If the \(\mathcal{R}\) value between two datasets is greater than 0.8, it is considered to be highly positively correlated \cite{cohen2013statistical}.

\subsection{Quantitative Analysis}

\paragraph{Overall Performance}

\begin{table}[t!]
\centering
\scriptsize  
\caption{Comparison of model results based on \textbf{human score}. The \colorbox{wkred}{first} and \colorbox{wkblue}{second} highest score are marked in red and blue. JN: just now, OD: one day, FD: few days, OM: one month, FM: few months, OY: one year, FY: few years, SD: several decades.}
\label{table:exp1_human}
\vskip 0.15in
\vspace{-2mm}
\setlength{\tabcolsep}{1.45mm}
\begin{tabular}{l|c|cccc cccc}
\toprule
Models & Overall & JN & OD & FD & OM & FM & OY & FY & SD  \\
\midrule
\multicolumn{10}{c}{\textbf{Easy}}\\
\midrule
ChatGLM3-6B & 2.7 & 2.8 & \best{7.2} & 2.7 & 1.8 & 2.8 & 0.0 & 1.0 & 3.5  \\
LLAMA3-8B & 5.6 & \second{6.0} & 4.2 & 6.1 & 4.6 & \second{6.0} & \second{5.6} & \second{4.7} & \best{8.0} \\
ChatGLM3-Turbo & 3.9 & 4.5 & 5.6 & 3.5 & 3.0 & 2.0 & 4.2 & 3.3 & 5.0 \\
GPT-3.5-turbo & 5.2 & 5.4 & 4.6 & \best{6.6} & \second{5.3} & 5.8 & 4.6 & 4.1 & 5.6 \\
GPT-4 & \second{5.8} & 5.0 & \second{6.6} & 6.1 & 5.0 & 5.8 & 6.0 & 4.0 & \second{7.8} \\
\texttt{\textbf{Echo}} (Ours) & \best{6.7} & \best{6.2} & 6.4 & \second{6.4} & \best{8.0} & \best{6.8} & \best{6.6} & \best{6.1} & 7.5 \\

\midrule
Mean Value (Easy) & 5.0 & 5.0 & 5.8 & 5.2 & 4.5 & 4.8 & 4.5 & 3.9 & 6.2 \\
\midrule
\multicolumn{10}{c}{\textbf{Hard}}\\
\midrule

ChatGLM3-6B & 2.6 & 2.0 & 4.8 & 1.9 & 3.5 & 1.0 & 1.8 & 1.9 & 2.0  \\
LLAMA3-8B & \second{5.5} & \best{7.0} & 5.2 & 4.0 & \second{4.5} & \second{6.4} & 5.0 & \second{5.1} & \best{7.0} \\
ChatGLM3-Turbo & 3.5 & 2.7 & \best{6.2} & 2.4 & 1.8 & 4.9 & 4.3 & 3.2 & 3.0 \\
GPT-3.5-turbo & 4.0 & 5.7 & 4.4 & 2.9 & 1.8 & 6.1 & 4.8 & 3.0 & 3.0 \\
GPT-4 & 5.4 & 5.7 & 5.4 & \second{5.0} & 3.5 & 5.6 & \best{6.0} & \second{5.1} & \best{7.0} \\
\texttt{\textbf{Echo}} (Ours) & \best{5.9} & \second{5.9} & \second{6.0} & \best{6.1} & \best{5.5} & \best{6.9} & \second{5.8} & \best{5.7} & \second{5.6} \\

\midrule
Mean Value (Hard) & 4.5 & 4.8 & 5.3 & 3.7 & 3.4 & 5.1 & 4.6 & 4.0 & 4.6 \\
\bottomrule
\end{tabular}
\vspace{-2mm}
\end{table}

\begin{table}[t!]
\centering
\scriptsize
\caption{Comparison of model results based on \textbf{similarity metric}.}
\label{table:exp1_simi}
\vskip 0.15in
\vspace{-2mm}
\setlength{\tabcolsep}{1.1mm}
\begin{tabular}{l|c|cccc cccc}
\toprule
Models & Overall & JN & OD & FD & OM & FM & OY & FY & SD  \\

\midrule
\multicolumn{10}{c}{\textbf{Easy}}\\
\midrule
ChatGLM3-6B & 57.0 & 42.7 & 76.7 & 50.8 & 52.2 & 37.2 & 50.1 & 68.2 & 77.9 \\
LLAMA3-8B & 70.2 & 64.3 & \second{77.0} & 73.6 & 64.2 & 56.6 & 78.1 & 59.7 & \best{87.7} \\
ChatGLM3-Turbo & 58.3 & 45.6 & 68.0 & 66.2 & 43.5 & 40.4 & 57.4 & 63.0 & 81.9 \\
GPT-3.5-turbo & \second{74.8} & \second{68.2} & 66.5 & \best{90.4} & \second{74.8} & \second{56.7} & 74.9 & \second{85.8} & 81.5 \\
GPT-4 & 72.3 & 63.6 & 72.3 & 76.8 & 64.2 & 53.5 & \second{81.0} & 80.8 & 86.5 \\
\texttt{\textbf{Echo}} (Ours) & \best{84.0} & \best{85.6} & \best{77.1} & \second{86.2} & \best{81.5} & \best{79.4} & \best{86.7} & \best{88.1} & \second{87.1} \\

\midrule
Mean Value (Easy) & 68.4 & 60.7 & 71.9 & 73.0 & 62.4 & 53.0 & 70.4 & 73.3 & 82.8 \\
\midrule
\multicolumn{10}{c}{\textbf{Hard}}\\
\midrule

ChatGLM3-6B & 56.3 & 53.2 & \second{71.8} & 45.6 & 58.9 & 37.6 & 60.9 & 71.1 & 51.4 \\
LLAMA3-8B & 64.3 & 72.9 & 63.8 & 52.7 & 71.1 & 64.1 & 53.4 & 69.3 & 67.4 \\
ChatGLM3-Turbo & 52.5 & 55.2 & 59.2 & 44.3 & 46.9 & 55.0 & 40.6 & 66.9 & 51.8 \\
GPT-3.5-turbo & 65.1 & \second{76.1} & 63.3 & 52.0 & \second{71.8} & 63.8 & 53.8 & \second{74.0} & 66.1 \\
GPT-4 & \second{67.7} & 71.9 & 66.8 & \second{59.1} & 70.7 & \second{68.9} & \second{64.3} & 70.5 & \second{69.8} \\
\texttt{\textbf{Echo}} (Ours) & \best{74.5} & \best{82.1} & \best{73.1} & \best{69.9} & \best{78.6} & \best{71.7} & \best{73.3} & \best{75.4} & \best{72.0} \\

\midrule
Mean Value (Hard) & 63.4 & 68.6 & 66.3 & 53.9 & 66.4 & 60.2 & 57.7 & 71.2 & 63.1 \\

\bottomrule
\end{tabular}
\vspace{-2mm}
\end{table}

We present the results of LLMs on human scores and similarity metrics in Table \ref{table:exp1_human} and Table \ref{table:exp1_simi}, respectively. Our experimental analysis is provided from the following aspects: performance of several LLMs, human score and similarity metrics, comparisons across easy and hard levels, comparisons across different time spans, and consistency between human score and similarity metric.

\begin{figure*}[h]
\centering

\includegraphics[width=.9\linewidth]{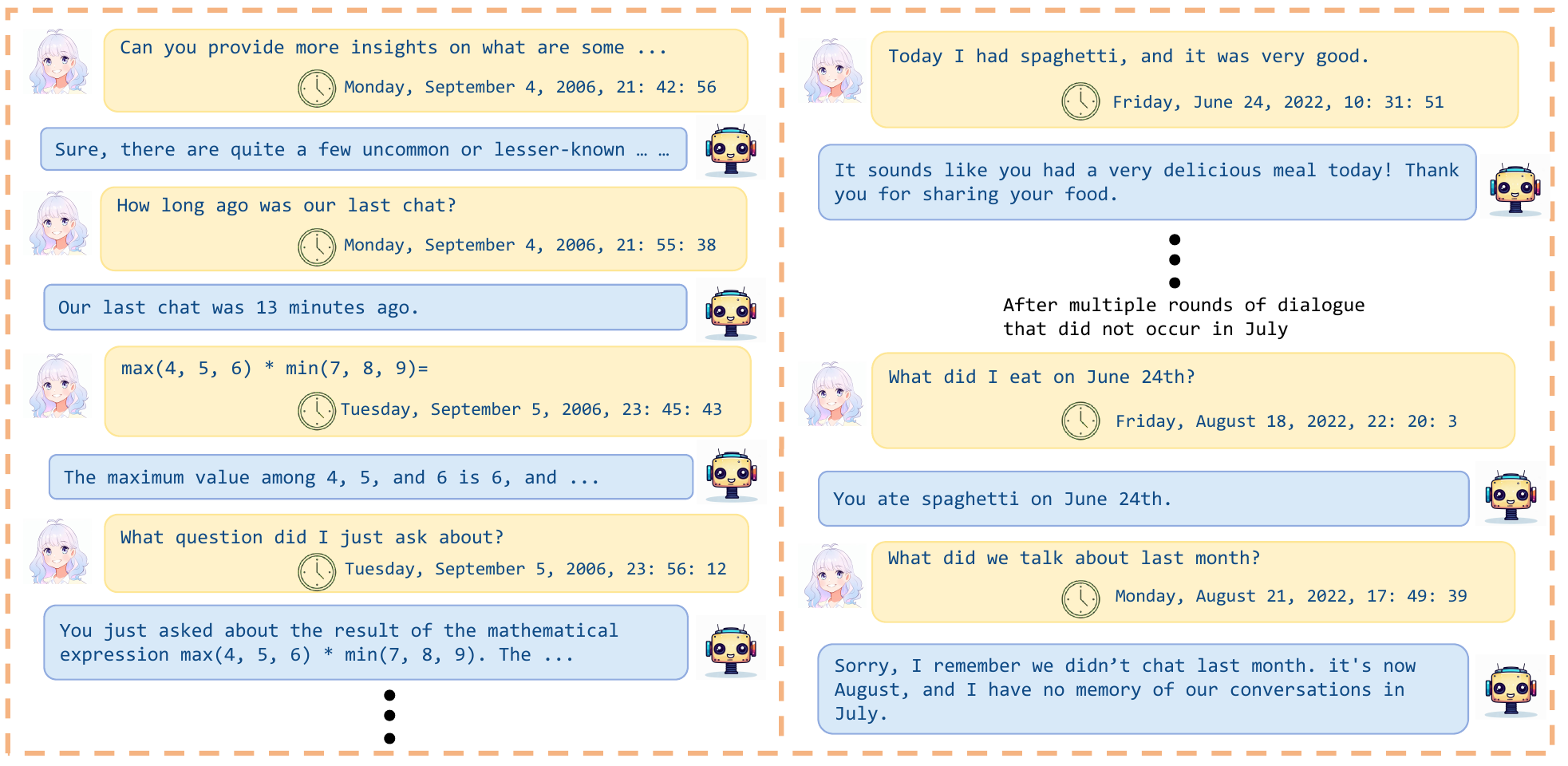} \\

\caption{Examples of complex episodic memory capability in the Echo.}
\label{fig:exp-demo1}

\end{figure*}

\textit{Performance of Several LLMs.}
It can be observed that our Echo model achieved the best performance in both Human Score and Similarity Metric. Specifically, it scored 6.7 and 5.9 in the easy and hard levels of Human Score, respectively, and 84.0 and 74.5 in the Similarity Metric. Over different time spans, Echo nearly obtained the best or second-best scores across all metrics, except for the easy level "One day" and "Several Decades" in Human Score. These results indicate that the Echo model excels in EM capability.
In contrast, among all models, the open-source ChatGLM3-6B performed the worst overall. As the base model of Echo, this indirectly demonstrates the effectiveness of the EM-Train data generated using the MADGF framework in enhancing a model's EM capability.
Moreover, GPT-4 also showed excellent comprehensive performance, achieving second-best scores in the easy level of Human Score and the hard level of Similarity Metric. In the hard level of Human Score, GPT-4 (5.4) narrowly trailed behind LLAMA3-8B (5.5), which had the second-best performance. In the easy level of Similarity Metric, GPT-4 (72.3) slightly lagged behind LLAMA3-8B (74.8), which also had the second-best performance.

\textit{Comparisons across Easy and Hard Levels.}
Most LLMs perform well on easy-level problems, but their performance drops at the hard level. Specifically, we found that all LLMs performed better overall at the Easy Level compared to the Hard Level. For instance, in Table \ref{table:exp1_human}, GPT-3.5-turbo scored 5.2 at the Easy Level, but only 4.0 at the Hard Level. Additionally, we observed that the average score for all models at the Hard Level in Table \ref{table:exp1_human} was 4.5, which is 0.5 lower than the Easy Level (5.0). Similar results were seen in Table \ref{table:exp1_simi}.

\textit{Comparisons across Different Time Spans.}
Models exhibit different performances across various time spans. Due to the inconsistency in model performance between the human score and similarity metric, we first consider the common performance under both metrics, then focus on the analysis based primarily on the human score. 
On the easy time span, we found that models perform better on "several decades" questions. This is because the mean values for "several decades" are the highest for all models according to both metrics (6.2 for human score and 82.8 for similarity metric). Meanwhile, we observed that models perform poorly on certain time spans (few days, few months, few years), which may be attributed to the difficulty models have in understanding these temporal concepts.

\begin{figure*}[h]
\centering

\includegraphics[width=.9\linewidth]{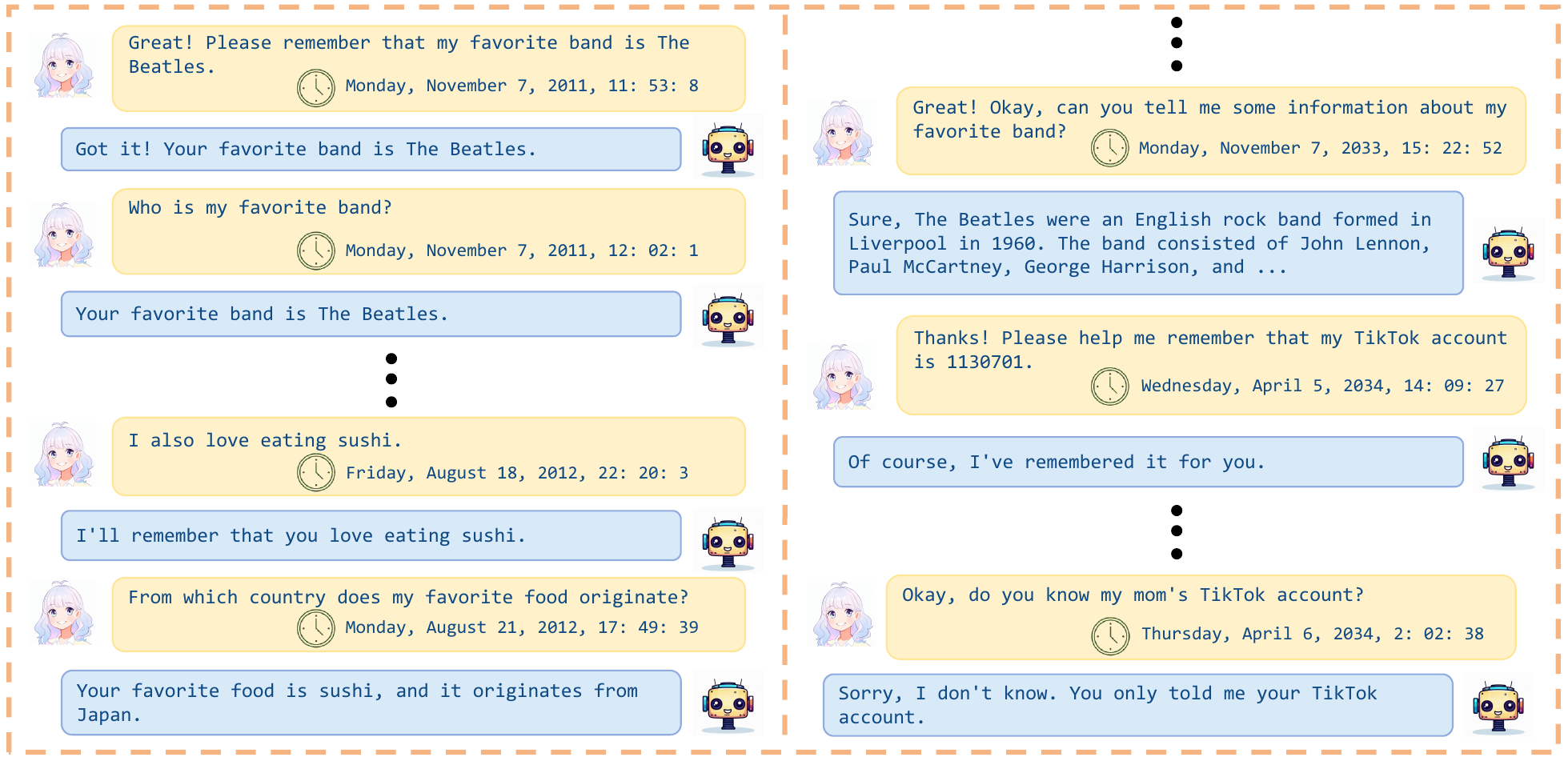} \\

\caption{Examples of episodic memory ability without temporal information in the Echo.}
\label{fig:exp-demo2}
\end{figure*}

\textit{Consistency Between Human Scores and Similarity Metrics.}
We consider calculating the Pearson correlation coefficient \( \mathcal{R} \) for two metrics to observe their correlation. The overall results of the Human Score (i.e., 2.7, 5.6, ..., 6.7) and the Similarity Metrics (i.e., 57.0, 70.2, ..., 84.0) are used to compute similarity at the Easy Level, and similarly for the Hard Level. Using Equation \ref{equ:pr}, we obtained Pearson correlation coefficients \( \mathcal{R} \) of 0.935 for the Easy Level and 0.842 for the Hard Level, both greater than 0.8. Therefore, the results of the two metrics can be considered highly positively correlated. Given that the Human Score requires expensive manual evaluation, Similarity Metrics can be considered as a cost-effective alternative for subsequent model evaluations.
Additionally, we found inconsistencies in the performance evaluation of models by Human Score and Similarity Metrics across different time spans. We calculated the Pearson correlation coefficient \( \mathcal{R} \) between the mean values of the two metrics over different time spans, resulting in coefficients of 0.429 (moderate correlation) and 0.003 (very weak correlation) for the Easy Level and Hard Level, respectively.
Our hypothesis is that the EM-Test may have insufficient data points for each time span, leading to inadequate statistical samples. This insufficiency results in inconsistent outcomes between Human Scores and Similarity Metrics.
\vspace{-2mm}

\paragraph{Performance in Episodic Memory Without Temporal Information}

We tested the EM capability of the models without considering time information, as shown in Table \ref{table:exp2}. We found that the models perform similarly whether or not time information is considered. Our Echo model and GPT-4 still performed well, achieving the first and second highest scores, respectively; while ChatGLM3-6B continued to perform the worst. These results indicate that the dataset EM-Train, obtained using our MADGF framework, can effectively improve the EM capability of models even when time information is not considered. In addition, we provide extended experiments regarding the model's temporal awareness and reasoning capability in the appendix for further analysis.

\begin{table}[t!]
\centering
\scriptsize
\caption{Comparison of performance in episodic memory without temporal information.}
\label{table:exp2}
\vskip 0.15in
\vspace{-2mm}
\setlength{\tabcolsep}{1.1mm}
\begin{tabular}{l|cc}
\toprule
Models & Easy & Hard   \\

\midrule

ChatGLM3-6B & 79.9 & 66.9 \\
LLAMA3-8B & 89.0 & 73.3  \\
ChatGLM3-Turbo & 84.3 & 67.0 \\
GPT-3.5-turbo & 90.0 & 77.8\\
GPT-4 & \second{90.4} & \second{78.3}\\
\texttt{\textbf{Echo}} (Ours) & \best{96.0} & \best{81.8}\\

\midrule
Mean Value & 88.3 & 74.2\\
\bottomrule
\end{tabular}
\vspace{-2mm}
\end{table}

\subsection{Qualitative Analysis}

\paragraph{Analysis of Complex Episodic Memory Ability} 
We conducted experiments on the Echo model using real-life scenarios that require complex Episodic Memory abilities, as shown in Figure \ref{fig:exp-demo1}. Some dialogues unrelated to episodic memory have been omitted using vertical ellipses. These dialogues are intended to increase the challenge of improving the model's Episodic Memory ability in long texts. Additionally, some content details unrelated to episodic memory skills have also been omitted. To test the model's performance over longer time spans, all time information was manually provided.

The test dialogue on the left side of Figure \ref{fig:exp-demo1} demonstrates that the model can accurately recall recent conversation content and timing, indicating its ability to understand time and associate it with events, which is a sign of Episodic Memory capability. In the test questions on the right side, the model shows even stronger Episodic Memory ability by recalling what human characters ate on a specific day from lengthy historical records, and understanding and judging whether conversations took place during a certain period.

\paragraph{Analysis of Episodic Memory Ability Without Temporal Information}

We tested Echo using questions that do not require considering time information for responses, as shown in Figure \ref{fig:exp-demo2}. From the test dialogue, it is clear that Echo can accurately recall the human character's favorite band and food, and provide relevant information even after multiple rounds of dialogue. Additionally, in the final round of test questions, Echo did not confuse any content that we had not actually told it, avoiding the hallucination issue. This problem often occurs when conversing with other LLMs.

\section{Conclusion}

In this paper, we investigate the episodic memory capabilities of LLMs. We propose an innovative Multi-agent data generation framework to collect high-quality, context-rich fine-tuning data, named EM-Train, which we used to further train the Echo model. We innovatively introduce time information into the training paradigm of LLMs. We also develop a multi-round dialogue test set, EM-Test, to evaluate the episodic memory capabilities of LLMs. Experimental results show that EM-Train significantly improves the Episodic Memory of LLMs. The experiments also verify that LLMs can gain time perception and reasoning abilities by incorporating time information into their training paradigms. Furthermore, qualitative experimental analysis indicates that Echo exhibits some human-like episodic memory capabilities. Our research provides a preliminary exploration of complex episodic memory capabilities with temporal information for LLMs.

\nocite{langley00}

\bibliography{main}

\begin{thebibliography}{34}
\providecommand{\natexlab}[1]{#1}
\providecommand{\url}[1]{\texttt{#1}}
\expandafter\ifx\csname urlstyle\endcsname\relax
  \providecommand{\doi}[1]{doi: #1}\else
  \providecommand{\doi}{doi: \begingroup \urlstyle{rm}\Url}\fi

\bibitem[B{\"a}rmann et~al.(2024)B{\"a}rmann, DeChant, Plewnia, Peller-Konrad, Bauer, Asfour, and Waibel]{barmann2024episodic}
B{\"a}rmann, L., DeChant, C., Plewnia, J., Peller-Konrad, F., Bauer, D., Asfour, T., and Waibel, A.
\newblock Episodic memory verbalization using hierarchical representations of life-long robot experience.
\newblock \emph{arXiv preprint arXiv:2409.17702}, 2024.

\bibitem[Cohen(2013)]{cohen2013statistical}
Cohen, J.
\newblock \emph{Statistical power analysis for the behavioral sciences}.
\newblock routledge, 2013.

\bibitem[Dan et~al.(2023)Dan, Lei, Gu, Li, Yin, Lin, Ye, Tie, Zhou, Wang, et~al.]{dan2023educhat}
Dan, Y., Lei, Z., Gu, Y., Li, Y., Yin, J., Lin, J., Ye, L., Tie, Z., Zhou, Y., Wang, Y., et~al.
\newblock Educhat: A large-scale language model-based chatbot system for intelligent education.
\newblock \emph{arXiv preprint arXiv:2308.02773}, 2023.

\bibitem[Ding et~al.(2023)Ding, Chen, Xu, Qin, Zheng, Hu, Liu, Sun, and Zhou]{ding2023enhancing}
Ding, N., Chen, Y., Xu, B., Qin, Y., Zheng, Z., Hu, S., Liu, Z., Sun, M., and Zhou, B.
\newblock Enhancing chat language models by scaling high-quality instructional conversations.
\newblock \emph{arXiv preprint arXiv:2305.14233}, 2023.

\bibitem[Dubey et~al.(2024)Dubey, Jauhri, Pandey, Kadian, Al-Dahle, Letman, Mathur, Schelten, Yang, Fan, et~al.]{dubey2024llama}
Dubey, A., Jauhri, A., Pandey, A., Kadian, A., Al-Dahle, A., Letman, A., Mathur, A., Schelten, A., Yang, A., Fan, A., et~al.
\newblock The llama 3 herd of models.
\newblock \emph{arXiv preprint arXiv:2407.21783}, 2024.

\bibitem[Fountas et~al.(2024)Fountas, Benfeghoul, Oomerjee, Christopoulou, Lampouras, Bou-Ammar, and Wang]{fountas2024human}
Fountas, Z., Benfeghoul, M.~A., Oomerjee, A., Christopoulou, F., Lampouras, G., Bou-Ammar, H., and Wang, J.
\newblock Human-like episodic memory for infinite context llms.
\newblock \emph{arXiv preprint arXiv:2407.09450}, 2024.

\bibitem[Gao \& Zhang(2024)Gao and Zhang]{gao2024memory}
Gao, H. and Zhang, Y.
\newblock Memory sharing for large language model based agents.
\newblock \emph{arXiv preprint arXiv:2404.09982}, 2024.

\bibitem[GLM et~al.(2024)GLM, Zeng, Xu, Wang, Zhang, Yin, Rojas, Feng, Zhao, Lai, Yu, Wang, Sun, Zhang, Cheng, Gui, Tang, Zhang, Li, Zhao, Wu, Zhong, Liu, Huang, Zhang, Zheng, Lu, Duan, Zhang, Cao, Yang, Tam, Zhao, Liu, Xia, Zhang, Gu, Lv, Liu, Liu, Yang, Song, Zhang, An, Xu, Niu, Yang, Li, Bai, Dong, Qi, Wang, Yang, Du, Hou, and Wang]{glm2024chatglm}
GLM, T., Zeng, A., Xu, B., Wang, B., Zhang, C., Yin, D., Rojas, D., Feng, G., Zhao, H., Lai, H., Yu, H., Wang, H., Sun, J., Zhang, J., Cheng, J., Gui, J., Tang, J., Zhang, J., Li, J., Zhao, L., Wu, L., Zhong, L., Liu, M., Huang, M., Zhang, P., Zheng, Q., Lu, R., Duan, S., Zhang, S., Cao, S., Yang, S., Tam, W.~L., Zhao, W., Liu, X., Xia, X., Zhang, X., Gu, X., Lv, X., Liu, X., Liu, X., Yang, X., Song, X., Zhang, X., An, Y., Xu, Y., Niu, Y., Yang, Y., Li, Y., Bai, Y., Dong, Y., Qi, Z., Wang, Z., Yang, Z., Du, Z., Hou, Z., and Wang, Z.
\newblock Chatglm: A family of large language models from glm-130b to glm-4 all tools, 2024.

\bibitem[Hu et~al.(2023)Hu, Fu, Du, Luo, Zhao, and Zhao]{hu2023chatdb}
Hu, C., Fu, J., Du, C., Luo, S., Zhao, J., and Zhao, H.
\newblock Chatdb: Augmenting llms with databases as their symbolic memory.
\newblock \emph{arXiv preprint arXiv:2306.03901}, 2023.

\bibitem[Ke et~al.(2024)Ke, Tong, Chen, and Peng]{ke2024exploring}
Ke, L., Tong, S., Chen, P., and Peng, K.
\newblock Exploring the frontiers of llms in psychological applications: A comprehensive review.
\newblock \emph{arXiv preprint arXiv:2401.01519}, 2024.

\bibitem[Langley(2000)]{langley00}
Langley, P.
\newblock Crafting papers on machine learning.
\newblock In Langley, P. (ed.), \emph{Proceedings of the 17th International Conference on Machine Learning (ICML 2000)}, pp.\  1207--1216, Stanford, CA, 2000. Morgan Kaufmann.

\bibitem[Li et~al.(2023)Li, Hammoud, Itani, Khizbullin, and Ghanem]{li2023camel}
Li, G., Hammoud, H., Itani, H., Khizbullin, D., and Ghanem, B.
\newblock Camel: Communicative agents for" mind" exploration of large language model society.
\newblock \emph{Advances in Neural Information Processing Systems}, 36:\penalty0 51991--52008, 2023.

\bibitem[Liu et~al.(2023)Liu, Hu, Zhou, Ding, Li, Zeng, He, Chen, Jiang, Zhou, et~al.]{liu2023mathematical}
Liu, W., Hu, H., Zhou, J., Ding, Y., Li, J., Zeng, J., He, M., Chen, Q., Jiang, B., Zhou, A., et~al.
\newblock Mathematical language models: A survey.
\newblock \emph{arXiv preprint arXiv:2312.07622}, 2023.

\bibitem[Luo et~al.(2023)Luo, Sun, Xu, Zhao, Lou, Tao, Geng, Lin, Chen, and Zhang]{luo2023wizardmath}
Luo, H., Sun, Q., Xu, C., Zhao, P., Lou, J., Tao, C., Geng, X., Lin, Q., Chen, S., and Zhang, D.
\newblock Wizardmath: Empowering mathematical reasoning for large language models via reinforced evol-instruct.
\newblock \emph{arXiv preprint arXiv:2308.09583}, 2023.

\bibitem[Moscovitch et~al.(2016)Moscovitch, Cabeza, Winocur, and Nadel]{moscovitch2016episodic}
Moscovitch, M., Cabeza, R., Winocur, G., and Nadel, L.
\newblock Episodic memory and beyond: the hippocampus and neocortex in transformation.
\newblock \emph{Annual review of psychology}, 67\penalty0 (1):\penalty0 105--134, 2016.

\bibitem[Naveed et~al.(2023)Naveed, Khan, Qiu, Saqib, Anwar, Usman, Akhtar, Barnes, and Mian]{naveed2023comprehensive}
Naveed, H., Khan, A.~U., Qiu, S., Saqib, M., Anwar, S., Usman, M., Akhtar, N., Barnes, N., and Mian, A.
\newblock A comprehensive overview of large language models.
\newblock \emph{arXiv preprint arXiv:2307.06435}, 2023.

\bibitem[OpenAI(2023)]{openai2023gpt4}
OpenAI.
\newblock Gpt-4 technical report, 2023.

\bibitem[Packer et~al.(2023)Packer, Wooders, Lin, Fang, Patil, Stoica, and Gonzalez]{packer2023memgpt}
Packer, C., Wooders, S., Lin, K., Fang, V., Patil, S.~G., Stoica, I., and Gonzalez, J.~E.
\newblock Memgpt: Towards llms as operating systems.
\newblock \emph{arXiv preprint arXiv:2310.08560}, 2023.

\bibitem[Qin et~al.(2024)Qin, Hu, Lin, Chen, Ding, Cui, Zeng, Huang, Xiao, Han, Fung, Su, Wang, Qian, Tian, Zhu, Liang, Shen, Xu, Zhang, Ye, Li, Tang, Yi, Zhu, Dai, Yan, Cong, Lu, Zhao, Huang, Yan, Han, Sun, Li, Phang, Yang, Wu, Ji, Liu, and Sun]{qin2024toollearningfoundationmodels}
Qin, Y., Hu, S., Lin, Y., Chen, W., Ding, N., Cui, G., Zeng, Z., Huang, Y., Xiao, C., Han, C., Fung, Y.~R., Su, Y., Wang, H., Qian, C., Tian, R., Zhu, K., Liang, S., Shen, X., Xu, B., Zhang, Z., Ye, Y., Li, B., Tang, Z., Yi, J., Zhu, Y., Dai, Z., Yan, L., Cong, X., Lu, Y., Zhao, W., Huang, Y., Yan, J., Han, X., Sun, X., Li, D., Phang, J., Yang, C., Wu, T., Ji, H., Liu, Z., and Sun, M.
\newblock Tool learning with foundation models, 2024.
\newblock URL \url{https://arxiv.org/abs/2304.08354}.

\bibitem[Schacter(2012)]{schacter2012constructive}
Schacter, D.~L.
\newblock Constructive memory: past and future.
\newblock \emph{Dialogues in clinical neuroscience}, 14\penalty0 (1):\penalty0 7--18, 2012.

\bibitem[Spens \& Burgess(2024)Spens and Burgess]{spens2024generative}
Spens, E. and Burgess, N.
\newblock A generative model of memory construction and consolidation.
\newblock \emph{Nature Human Behaviour}, 8\penalty0 (3):\penalty0 526--543, 2024.

\bibitem[Sprott(1933)]{sprott1933remembering}
Sprott, W.
\newblock Remembering: A study in experimental and social psychology., 1933.

\bibitem[Squire(2004)]{squire2004memory}
Squire, L.~R.
\newblock Memory systems of the brain: a brief history and current perspective.
\newblock \emph{Neurobiology of learning and memory}, 82\penalty0 (3):\penalty0 171--177, 2004.

\bibitem[Tan et~al.(2023)Tan, Ng, and Bing]{tan2023towards}
Tan, Q., Ng, H.~T., and Bing, L.
\newblock Towards benchmarking and improving the temporal reasoning capability of large language models.
\newblock \emph{arXiv preprint arXiv:2306.08952}, 2023.

\bibitem[Tulving(1972)]{tulving1972episodic}
Tulving, E.
\newblock Episodic and semantic memory.
\newblock \emph{Organization of memory/Academic Press}, 1972.

\bibitem[Tulving(1983)]{tulving1983elements}
Tulving, E.
\newblock Elements of episodic memory, 1983.

\bibitem[Wang et~al.(2022)Wang, Kordi, Mishra, Liu, Smith, Khashabi, and Hajishirzi]{wang2022self}
Wang, Y., Kordi, Y., Mishra, S., Liu, A., Smith, N.~A., Khashabi, D., and Hajishirzi, H.
\newblock Self-instruct: Aligning language models with self-generated instructions.
\newblock \emph{arXiv preprint arXiv:2212.10560}, 2022.

\bibitem[Wang et~al.(2023)Wang, Peng, Que, Liu, Zhou, Wu, Guo, Gan, Ni, Yang, et~al.]{wang2023rolellm}
Wang, Z.~M., Peng, Z., Que, H., Liu, J., Zhou, W., Wu, Y., Guo, H., Gan, R., Ni, Z., Yang, J., et~al.
\newblock Rolellm: Benchmarking, eliciting, and enhancing role-playing abilities of large language models.
\newblock \emph{arXiv preprint arXiv:2310.00746}, 2023.

\bibitem[Xu et~al.(2023)Xu, Sun, Zheng, Geng, Zhao, Feng, Tao, and Jiang]{xu2023wizardlm}
Xu, C., Sun, Q., Zheng, K., Geng, X., Zhao, P., Feng, J., Tao, C., and Jiang, D.
\newblock Wizardlm: Empowering large language models to follow complex instructions.
\newblock \emph{arXiv preprint arXiv:2304.12244}, 2023.

\bibitem[Yang et~al.(2024)Yang, Yang, Hui, Zheng, Yu, Zhou, Li, Li, Liu, Huang, et~al.]{yang2024qwen2}
Yang, A., Yang, B., Hui, B., Zheng, B., Yu, B., Zhou, C., Li, C., Li, C., Liu, D., Huang, F., et~al.
\newblock Qwen2 technical report.
\newblock \emph{arXiv preprint arXiv:2407.10671}, 2024.

\bibitem[Zhang et~al.(2023)Zhang, Chen, Liu, Liao, Gong, Yu, Li, and Wang]{zhang2023unifying}
Zhang, Z., Chen, C., Liu, B., Liao, C., Gong, Z., Yu, H., Li, J., and Wang, R.
\newblock Unifying the perspectives of nlp and software engineering: A survey on language models for code.
\newblock \emph{arXiv preprint arXiv:2311.07989}, 2023.

\bibitem[Zhao et~al.(2024)Zhao, Ren, Hessel, Cardie, Choi, and Deng]{zhao2024wildchat}
Zhao, W., Ren, X., Hessel, J., Cardie, C., Choi, Y., and Deng, Y.
\newblock Wildchat: 1m chatgpt interaction logs in the wild.
\newblock \emph{arXiv preprint arXiv:2405.01470}, 2024.

\bibitem[Zhao et~al.(2023)Zhao, Zhou, Li, Tang, Wang, Hou, Min, Zhang, Zhang, Dong, et~al.]{zhao2023survey}
Zhao, W.~X., Zhou, K., Li, J., Tang, T., Wang, X., Hou, Y., Min, Y., Zhang, B., Zhang, J., Dong, Z., et~al.
\newblock A survey of large language models.
\newblock \emph{arXiv preprint arXiv:2303.18223}, 2023.

\bibitem[Zhong et~al.(2024)Zhong, Guo, Gao, Ye, and Wang]{zhong2024memorybank}
Zhong, W., Guo, L., Gao, Q., Ye, H., and Wang, Y.
\newblock Memorybank: Enhancing large language models with long-term memory.
\newblock In \emph{Proceedings of the AAAI Conference on Artificial Intelligence}, volume~38, pp.\  19724--19731, 2024.

\end{thebibliography}
\bibliographystyle{icml2025}

\newpage
\appendix
\onecolumn
\section*{Appendix}

\appendix

\section{Implementation Details of MADGF}

\subsection{Plots used in Human Prompt Template}
\label{sec:plotsforhuman}

Table \ref{tab:humanplots} presents the plots used in the human prompt template. It includes 20 plots, with the numbers of \colorbox{wkblue}{true episodic memories} and  \colorbox{wkred}{hallucinatory episodic memories} marked in blue and red, respectively. The final plot is fixed as "say goodbye" to guide the conclusion of the conversation.

\begin{table}[h!]
\centering
\caption{Example of plots for human prompt.}
\begin{tabular}{>{\centering\arraybackslash}p{0.05\textwidth} p{0.9\textwidth}}
\toprule
\textbf{No.} & \textbf{Plot Description} \\
\midrule
1 & Ask what day of the week it is today \\
\second{2} & Request AI to inform you of the current date and time \\
\second{3} & Ask if we talked the day before yesterday, and if AI answers yes, then ask what topic we discussed \\
4 & Ask a question about earth science \\
\second{5} & Ask AI for its name and call it by that name instead of AI from now on \\
\second{6} & Ask AI to remember your fitness plan \\
\second{7} & Ask what day the next working day is \\
\best{8} & Ask a piece of information you haven’t told AI before: the date you first attended an online course \\
9 & Ask AI how it is feeling today \\
\best{10} & Ask a piece of information you haven’t told AI before: your cherished books \\
11 & Inquire about AI’s perspective on the development of artificial intelligence \\
\second{12} & Ask AI to remember your grandfather’s favorite news source \\
\second{13} & Ask AI if it remembers your fitness plan \\
14 & Ask a career-related question \\
\second{15} & Ask AI if it remembers your grandfather’s favorite news source, and if it does, ask when you shared it \\
\second{16} & Ask what day the last working day was \\
17 & Ask a simple physics question \\
\best{18} & Ask a piece of information you haven’t told AI before: the date of your first marathon completion \\
\best{19} & Ask a piece of information you haven’t told AI before: your private collection inventory \\
20 & Say goodbye \\
\bottomrule
\end{tabular}
\label{tab:humanplots}
\end{table}

\subsection{Hallucinatory Plots used in Assistant Prompt Template}
\label{sec:hplotsforassist}

Table \ref{tab:assistplots1} provides an example of hallucinatory plots used in the assistant prompt template, aimed at guiding the assistant to avoid hallucination issues. The example includes four memories that did not occur in actual conversations, corresponding to the plots marked in red (\colorbox{wkred}{8, 10, 18}, and \colorbox{wkred}{19}) in Table \ref{tab:humanplots}.


\begin{table}[h!]
\centering
\caption{Example of hallucinatory plots for assistant prompt.}
\begin{tabular}{>{\centering\arraybackslash}p{0.05\textwidth} >{\centering\arraybackslash}p{0.4\textwidth}}
\toprule
\textbf{No.} & \textbf{Noteworthy Hallucinatory Plots} \\
\midrule
1 & first day attending an online course \\
2 & cherished books \\
3 & first marathon completion date \\
4 & private collection inventory \\
\bottomrule
\end{tabular}
\label{tab:assistplots1}
\end{table}

\subsection{Common Plots used in Assistant Prompt Template}
\label{sec:cplotsforassist}

The common plots designed to prompt the AI assistant to proactively seek relevant information in a human-like manner. One example is "name, old, hobby, gender".



\section{Extended Experiments on Temporal Awareness and Reasoning Capability of the Model}

To enhance and evaluate the temporal awareness and reasoning capabilities of the model, we have developed temporally aware and reasoning-enhanced training and testing datasets. We then conducted both quantitative and qualitative experimental analyses of Echo.
\vspace{-2mm}

\subsection{Temporal Reasoning Dataset}

\paragraph{Training Dataset}
We improved upon a portion of the training set proposed by Tan et.al \cite{tan2023towards} to create a dataset that emphasizes temporal awareness and reasoning. In the work by Tan et.al \cite{tan2023towards}, the data were entirely synthesized programmatically, with questions being relatively simplistic, lacking inquiries about specific days of the week or recent dates. Utilizing both programming techniques and manual annotations, we constructed an 8K training dataset. The data format adheres to Echo's training paradigm of user-time-assistant, making it highly suitable for Echo model training. Table \ref{tab:time_train} provides examples from our training dataset, which includes various complex scenarios for temporal reasoning questions, aiding in developing Echo's robust temporal awareness and reasoning skills after training.
\vspace{-3mm}

\begin{table}[h!]
\centering
\caption{Examples of temporal reasoning dataset.}
\begin{tabular}{>{\centering\arraybackslash}p{0.05\textwidth} >{\raggedright\arraybackslash}p{0.7\textwidth}}
\toprule
\multicolumn{2}{c}{\textbf{Training Dataset}}\\
\midrule
\textbf{No.} & \multicolumn{1}{c}{\textbf{Question, Time, Answer}} \\
\midrule
1 & \makecell[l]{
    \textbf{Question}: What is the time 10 years and 6 months after June 1856? \\
    \textbf{Time}: Monday, September 9, 2024, 16:03:45 \\
    \textbf{Answer}: The time is Dec, 1866
} \\
\hline
2 & \makecell[l]{
    \textbf{Question}: What was the date a year ago today? \\
    \textbf{Time}: Friday, April 3, 2020, 5:04:46 \\
    \textbf{Answer}: Today is 2020-4-3, therefore a year ago today should be 2019-4-3.
} \\
\hline
3 & \makecell[l]{
    \textbf{Question}: What day of the week was yesterday? \\
    \textbf{Time}: Friday, April 23, 2049, 23:30:07 \\
    \textbf{Answer}: Yesterday was Thursday.
} \\
\hline
4 & \makecell[l]{
    \textbf{Question}: Will the day after tomorrow be a weekend? \\
    \textbf{Time}: Saturday, July 3, 2027, 19:17:33 \\
    \textbf{Answer}: No, the day after tomorrow will be Monday, July 5th, 2027.
} \\
\hline
5 & \makecell[l]{
    \textbf{Question}:How can I use PHP to compare the existentialist philosophies of ...  \\
    \textbf{Time}: Tuesday, June 22nd, 2038, 11: 01: 39 \\
    \textbf{Answer}: As an AI assistant, I can provide guidance on how to approach the task, ...\\
    \textbf{Question}: How long ago was our last chat? \\
    \textbf{Time}: Tuesday, June 22nd, 2038, 11: 31: 7 \\
    \textbf{Answer}: Our last conversation was just now, 30 minutes ago, at 11:1:39.
} \\
\midrule
\multicolumn{2}{c}{\textbf{Evaluation Dataset}}\\
\midrule
\textbf{No.} & \multicolumn{1}{c}{\textbf{Question, Time, Answer, key word}} \\
\hline
1 & \makecell[l]{
    \textbf{Question}: 432 years ago today, which day was it? \\
    \textbf{Time}: Tuesday, September 3, 2013, 23: 42: 54 \\
    \textbf{Answer}: Today, 432 years ago, was 1581-9-3. \\
    \textbf{key word}: "1581", "9\textbar Sep\textbar September", "3"
} \\
\bottomrule
\end{tabular}
\vspace{-3mm}
\label{tab:time_train}
\end{table}

\paragraph{Evaluation Dataset}
We manually annotated a temporally aware and reasoning-enhanced evaluation dataset consisting of 292 instances, including 32 short-term (within one week) and 260 long-term test questions, as shown in Table \ref{tab:etime_test}. Each test question provides all possible keywords contained in the standard answer, allowing for accurate quantitative analysis of whether the model's output is correct through string matching.

\vspace{-3mm}

\begin{table}[h!]
\centering
\caption{Statistics of the evaluation dataset for temporal reasoning}
\label{tab:etime_test}
\vskip 0.15in
\vspace{-2mm}
\setlength{\tabcolsep}{1.1mm}
\begin{tabular}{llll}
\toprule
\textbf{Time Span} & \textbf{Short}  & \textbf{Long} & \textbf{Total} \\
\hline
Overall Number & 32 & 260 & 292 \\
\bottomrule
\end{tabular}
\vspace{-2mm}
\end{table}

\subsection{Quantitative Analysis of Temporal Perception and Reasoning Ability}

On the Evaluation Dataset for Temporal Reasoning, we conducted a quantitative analysis. As in Section \ref{sec:experiment}, we selected LLAMA3-8b \cite{dubey2024llama} and ChatGLM3-6B \cite{glm2024chatglm} for open-source models, and for closed-source models, we employed GPT-3.5-turbo \cite{openai2023gpt4}, GPT-4 \cite{openai2023gpt4}, and ChatGLM3-turbo \cite{glm2024chatglm} for evaluation and comparison. When calculating the metrics, we detected keywords within the models' responses.

As shown in Table \ref{table:exp3}, we found that our Echo model still performs the best, with time-aware and reasoning abilities exceeding 90 in both short-term (98.1) and long-term (94.6) scenarios. In contrast, ChatGLM3-6B performed very poorly, with time-aware and reasoning abilities below 10 in both short-term (9.4) and long-term (8.8) scenarios. This indicates that the EM-Train dataset significantly improves the time-aware and reasoning capabilities of the models. Additionally, we observed that GPT-4 achieved suboptimal performance on long-term tests, but did not achieve suboptimal performance on short-term tests. Upon examining the model's outputs, we noticed that GPT-4 tends to produce errors and hallucinations in short-term temporal reasoning. For example, the correct answer was "The date the day before yesterday was July 1st, 2023.", but GPT-4's output was "The day before yesterday would have been July 2, 2023.".
 

\begin{table}[h]
\centering
\footnotesize
\caption{Comparison of performance in time perception and reasoning.}
\label{table:exp3}
\vskip 0.15in
\vspace{-2mm}
\setlength{\tabcolsep}{1.1mm}
\begin{tabular}{l|cc}
\toprule
Models & Short-term & Long-term   \\

\midrule

ChatGLM3-6B & 9.4 & 8.8 \\
LLAMA3-8B & \second{90.6} & 65.0  \\
ChatGLM3-Turbo & 46.9 & 34.6 \\
GPT-3.5-turbo & 56.2 & 78.1\\
GPT-4 & 78.1 & \second{93.8}\\
\texttt{\textbf{Echo}} (Ours) & \best{98.1} & \best{94.6}\\

\midrule
Mean Value & 63.2 & 62.5\\
\bottomrule
\end{tabular}
\vspace{-2mm}
\end{table}

\subsection{Qualitative Analysis of Temporal Perception and Reasoning Ability}

\begin{figure*}[h]
\centering

\includegraphics[width=.9\linewidth]{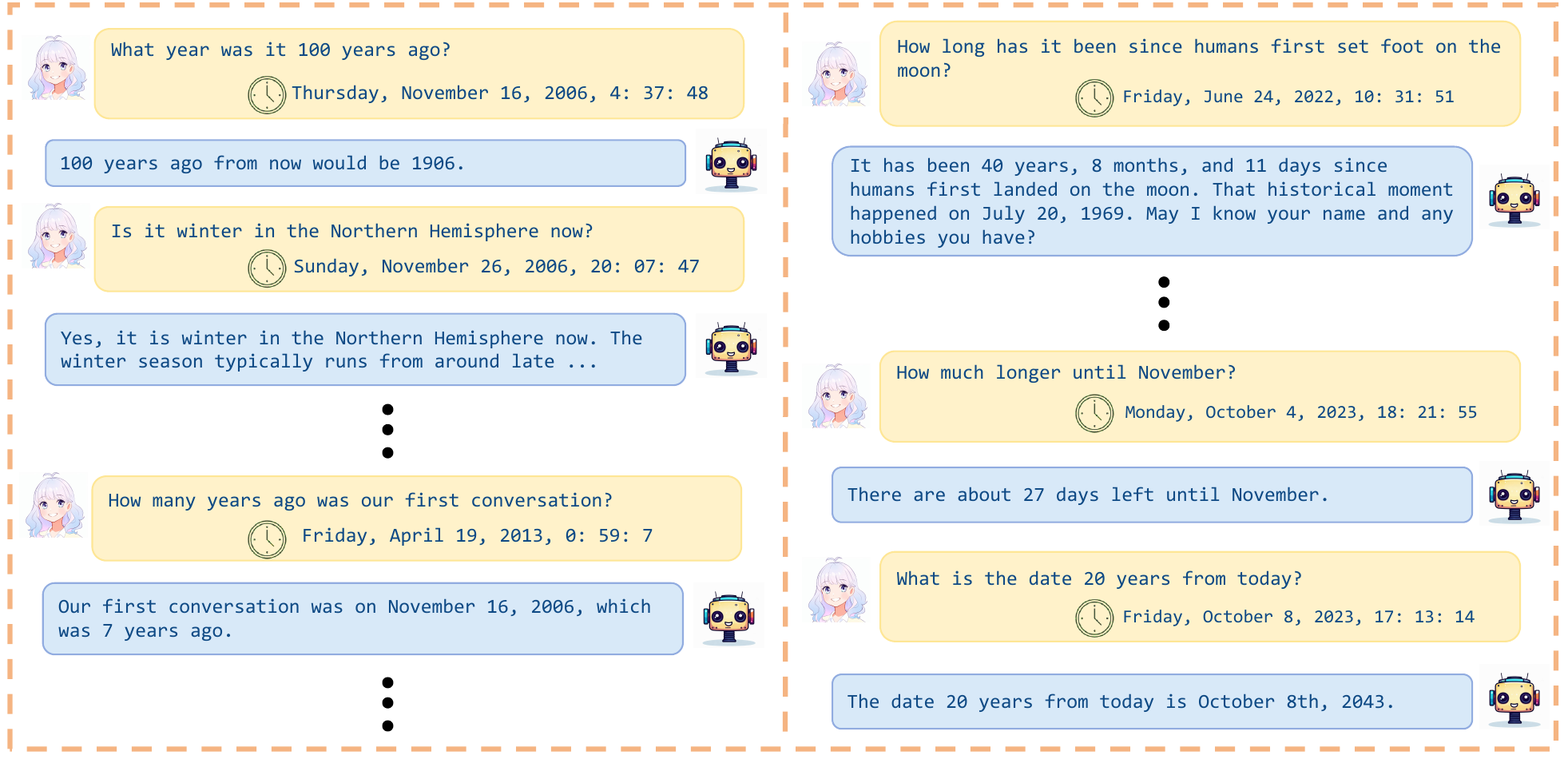} \\

\caption{Examples of time perception and reasoning ability in the Echo.}
\label{fig:exp-demo3}
\end{figure*}

We present the qualitative analysis results of Echo in Figure \ref{fig:exp-demo3}. It is evident that the model can perceive the current time and perform reasoning tasks, such as correctly answering questions about the current season and how many years have passed since the first moon landing. Additionally, the model can also perceive and reason about past and future times. For example, it accurately answered questions about what the date will be 100 years and 20 years from now, how long until November, and how much time has passed since the first chat session.

\end{document}